%% file: chen2022ral.tex
\documentclass[letterpaper, 10 pt, conference]{ieeeconf}
\IEEEoverridecommandlockouts    % This command is only needed if
\input{stachnisslab-latex}

\input{stachnisslab-math}

\usepackage{subfigure}
\usepackage{pifont}
\usepackage{amssymb}
\usepackage{graphicx}
\graphicspath{{pics/}}

\title{\LARGE \bf Automatic Labeling to Generate Training Data \\ for Online LiDAR-based Moving Object Segmentation}

\author{Xieyuanli\,Chen \and Benedikt\,Mersch \and Lucas\,Nunes \and  Rodrigo\,Marcuzzi \and Ignacio\,Vizzo \and Jens\,Behley \and Cyrill\,Stachniss% <-this % stops a space
	\thanks{All authors are with the University of Bonn, Germany.}%
	\thanks{This work has partially been funded by the Deutsche Forschungsgemeinschaft (DFG, German Research Foundation) under Germany's Excellence Strategy, EXC-2070 - 390732324 - PhenoRob, by the EC within Horizon Europe, grant agreemenet no.~101017008 (Harmony), and by the Chinese Scholarship Committee.
	}%
}

\begin{document}
	\maketitle
	\thispagestyle{empty}
	\pagestyle{empty}

%%%%%%%%%%%%%%%%%%%%%%%%%%%%%%%%%%%%%%%%%%%%%%%%%%%%%%%%%%%%%%%%%%%%%%%%%%%%%%%%
\begin{abstract}
  Understanding the scene is key for autonomously navigating vehicles and the ability to segment the surroundings online into moving and non-moving objects is a central ingredient for this task. Often, deep learning-based methods are used to perform moving object segmentation~(MOS). The performance of these networks, however, strongly depends on the diversity and amount of \emph{labeled} training data---information that may be costly to obtain. In this paper, we propose an automatic data labeling pipeline for 3D LiDAR data to save the extensive manual labeling effort and to improve the performance of existing learning-based MOS systems by automatically generating labeled training data. Our proposed approach achieves this by processing the data offline in batches. It first exploits an occupancy-based dynamic object removal to detect possible dynamic objects coarsely. Second, it extracts segments among the proposals and tracks them using a Kalman filter. Based on the tracked trajectories, it labels the actually moving objects such as driving cars and pedestrians as moving. In contrast, the non-moving objects, e.g., parked cars, lamps, roads, or buildings, are labeled as static. We show that this approach allows us to label LiDAR data highly effectively and compare our results to those of other label generation methods. We also train a deep neural network with our auto-generated labels and achieve similar performance compared to the one trained with manual labels on the same data---and an even better performance when using additional datasets with labels generated by our approach. Furthermore, we evaluate our method on multiple datasets using different sensors and our experiments indicate that our method can generate labels in diverse environments.
\end{abstract}

%%%%%%%%%%%%%%%%%%%%%%%%%%%%%%%%%%%%%%%%%%%%%%%%%%%%%%%%%%%%%%%%%%%%%%%%%%%%%%%%
\section{Introduction}
\label{sec:intro}

Moving object segmentation~(MOS) is a fundamental task for autonomous vehicles as it separates the \emph{actually moving} objects such as driving cars and pedestrians from static or non-moving objects such as buildings, parked cars, etc. This is an important processing step needed in many applications, such as predicting the future state of the surroundings~\cite{mersch2021corl}, collision avoidance~\cite{peters2021rss}, or robot path planning~\cite{kummerle2014jfr}. This knowledge can also improve and robustify pose estimation, sensor data registration, and simultaneous localization and mapping~(SLAM)~\cite{chen2019iros}. Thus, accurate and reliable MOS available at frame-rate is relevant for most autonomous mobile systems. 
	
\begin{figure}[t]
	\centering
	\includegraphics[width=0.85\linewidth]{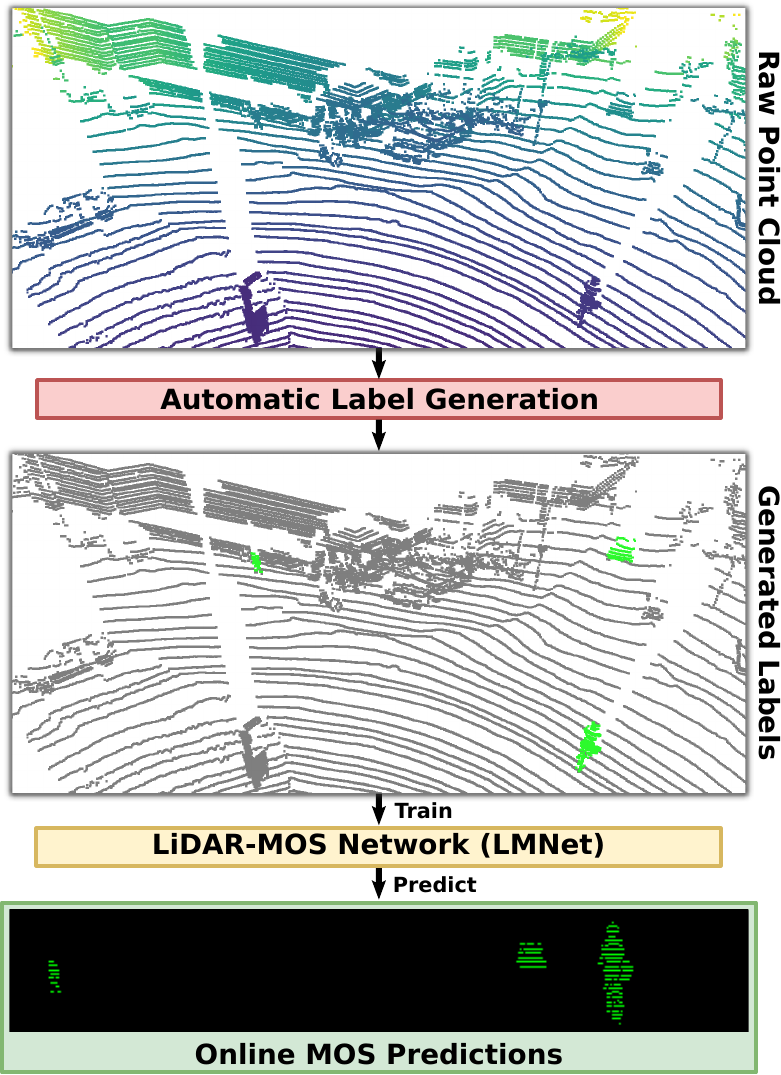}
	\caption{Moving object segmentation trained with the labels automatically generated by our approach. Our method can detect and segment the moving objects given sequential point cloud data and generate labels for LiDAR-based moving object segmentation. Based on the automatically generated labels, we re-train our LiDAR-MOS network, which can be deployed on an autonomous vehicle to perform LiDAR-MOS online in unseen environments.}
	\label{fig:motivation}
  \vspace{-0.5cm}
\end{figure}

3D~LiDAR-based moving object segmentation (LiDAR-MOS) is challenging due to the distance-dependent sparsity and uneven distribution of the range measurements. Our recent work~\cite{chen2021ral} proposes a deep neural network, called LMNet, to achieve LiDAR-MOS faster than sensor frame rate by exploiting sequential LiDAR range images. Such supervised learning-based methods rely on often manually labeled data, which is often limited in size and cannot easily be generated for new or unseen environments. Despite a large amount of publicly available LiDAR datasets nowadays~\cite{behley2021ijrr, caesar2020cvpr, geiger2013ijrr, liao2021arxiv}, labeling LiDAR data is still a tedious process and one of the bottlenecks in supervised learning.	Automatic label generation can alleviate this problem by exploiting the temporal-spatial dependence of the dataset and, compared to online operation, that the data can be processed in batches. For example, the labels at a specific timestamp can be easier determined when exploiting preceding \emph{and} succeeding scans compared to online MOS. 

The main contribution of this paper is a novel modularized approach to generate MOS labels in 3D LiDAR scans automatically. Our approach first exploits occupancy-based dynamic object removal techniques to detect possible dynamic objects coarsely. We then cluster the proposals into instances and track them using a Kalman filter. Based on the tracked trajectories, we label the actually moving objects (driving cars, pedestrians, etc.) as moving. In contrast, the non-moving objects, including parked cars, are labeled as static. Based on the labels automatically generated offline, we train the LiDAR-MOS network LMNet~\cite{chen2021ral}. Note that no manually labeled data or other sensor information is needed for training. Once having a sequence of LiDAR scans, our method can automatically generate MOS labels.
  
For evaluation, we automatically label LiDAR scans in the training sequences of the KITTI odometry dataset~\cite{geiger2013ijrr} using different methods and compare the quality of generated MOS labels using the LiDAR-MOS labels from SemanticKITTI~\cite{behley2019iccv, chen2021ral}. Next, we automatically generate more labels on additional data from KITTI, \ie, the KITTI road dataset, to train LMNet further. Compared to the network trained on manually labeled ground truth data, the evaluation results suggest that the network trained on labels automatically generated by our method achieves a similar performance, and is superior when using more automatically generated data from additional scans. Furthermore, our approach generalizes well to different, unseen environments, which we show for MOS on three different datasets. 

In sum, we make the following claims:
(i)~Compared to existing methods, our approach generates better labels for LiDAR-MOS.
(ii)~Based on our generated labels, a network achieves a similar performance compared to the same network trained with manual labels on the same data and better performance with additional automatically labeled training data.
(iii)~Our method generates effective labels for different LiDAR scanners and in different environments.

%%%%%%%%%%%%%%%%%%%%%%%%%%%%%%%%%%%%%%%%%%%%%%%%%%%%%%%%%%%%%%%%%%%%%%%%%%%%%%%%
\section{Related Work}
\label{sec:related}
	
	With the advent of deep neural networks, the task of moving object segmentation has achieved great success for image and video-based approaches~\cite{giraldo2020tpami, patil2020cvpr}.
	However, it is still challenging for the LiDAR-based approach due to the sparsity of the sensor data and the lack of publicly available large-scale datasets with point-wise moving and static labels. 
	Here, we concentrate on approaches LiDAR-based MOS and methods that can generate LiDAR-MOS labels.
	
	There are both geometric model-based and deep neural network-based methods in the literature to address the problem of \emph{online LiDAR-MOS}.
	For example, Yoon \etalcite{yoon2019cvr} detect dynamic objects in LiDAR scans exploiting geometric heuristics, \eg, the residual between LiDAR scans, free space checking, and region growing to find moving objects. Dewan \etalcite{dewan2016iros} propose a LiDAR-based scene flow method that estimates motion vectors for rigid bodies.
	
	Besides geometry-based approaches, there are also deep network-based methods~\cite{baur2021iccv, gojcic2021cvpr, liu2019cvpr}, which use generic end-to-end trainable models to learn local and global statistical relationships directly from data.
	Such scene flow methods usually estimate motion vectors between two consecutive scans, which may not differentiate between slowly moving objects and sensor noise. 
	Semantic segmentation can be seen as a relevant step towards moving object segmentation. 
	Based on the online semantic segmentation~\cite{milioto2019iros,cortinhal2020iv,li2021ral}, SuMa++~\cite{chen2019iros} exploits semantics to detect and filter out dynamic objects to improve the LiDAR SLAM performance.
	There are also several 3D point cloud-based semantic segmentation approaches~\cite{shi2020cvpr,thomas2019iccv}, which exploit sequential point clouds and predict moving objects.
	However, these methods require a large number of semantic labels, which are not always available and make it difficult to train and generalize to unseen data.
  Recently, we proposed a deep learning-based LiDAR-MOS approach~\cite{chen2021ral} exploiting short sequences of range images from a rotating 3D LiDAR sensor, which achieves accurate online performance on a novel LiDAR-MOS benchmark.
	Despite good performance and simplified binary labels, it still relies on manual labels and does not generalize well to different environments with varying appearances.   
	
	Unlike online LiDAR-MOS, label generation can be done \emph{offline}, which enables better performance in perception tasks, using sequential future and past observations.
	Multiple works have already been proposed to clean the point cloud map, which removes dynamic objects during the mapping procedure, resulting in a clean static map~\cite{arora2021ecmr, gehrung2017isprsannals, kim2020iros, lim2021ral, pagad2020icra, pomerleau2014icra, schauer2018ral, xiao2015jprs}. 
	There are methods~\cite{gehrung2017isprsannals, schauer2018ral} that use the time-consuming voxel ray casting and require accurately aligned poses to clean dense terrestrial laser scans.
	To alleviate the computational burden, visibility-based methods have been proposed~\cite{pomerleau2014icra, xiao2015jprs}. These types of methods associate a query point cloud to a map point within a narrow field of view, \eg, cone-shaped used by Pomerleau~\etalcite{pomerleau2014icra}.
	Kim~\etalcite{kim2020iros} propose a range image-based method, which exploits the consistency check between the query scan and the pre-built map to remove dynamic points. The authors use a multi-resolution false prediction reverting algorithm to refine the map.
	Pagad~\etalcite{pagad2020icra} propose an occupancy map-based method to remove dynamic points in LiDAR scans. They first build occupancy maps using object detection and then use the voxel traversal method to remove the moving objects.
	Recently, Lim~\etalcite{lim2021ral} propose a method to first remove dynamic objects by checking the occupancy of each sector of LiDAR scans and then revert the ground plane by region growth. In contrast, Arora \etalcite{arora2021ecmr} first segment out the ground plane and then remove the ``ghost effect'' caused by the moving object during mapping.
	Even though such map cleaning methods can distinguish the moving objects from the static map, they usually remove not only dynamic objects but also static parts due to noisy poses or occlusions.

\begin{figure*}[t]
\centering
	\vspace{+0.2cm}
	\includegraphics[width=0.9\linewidth]{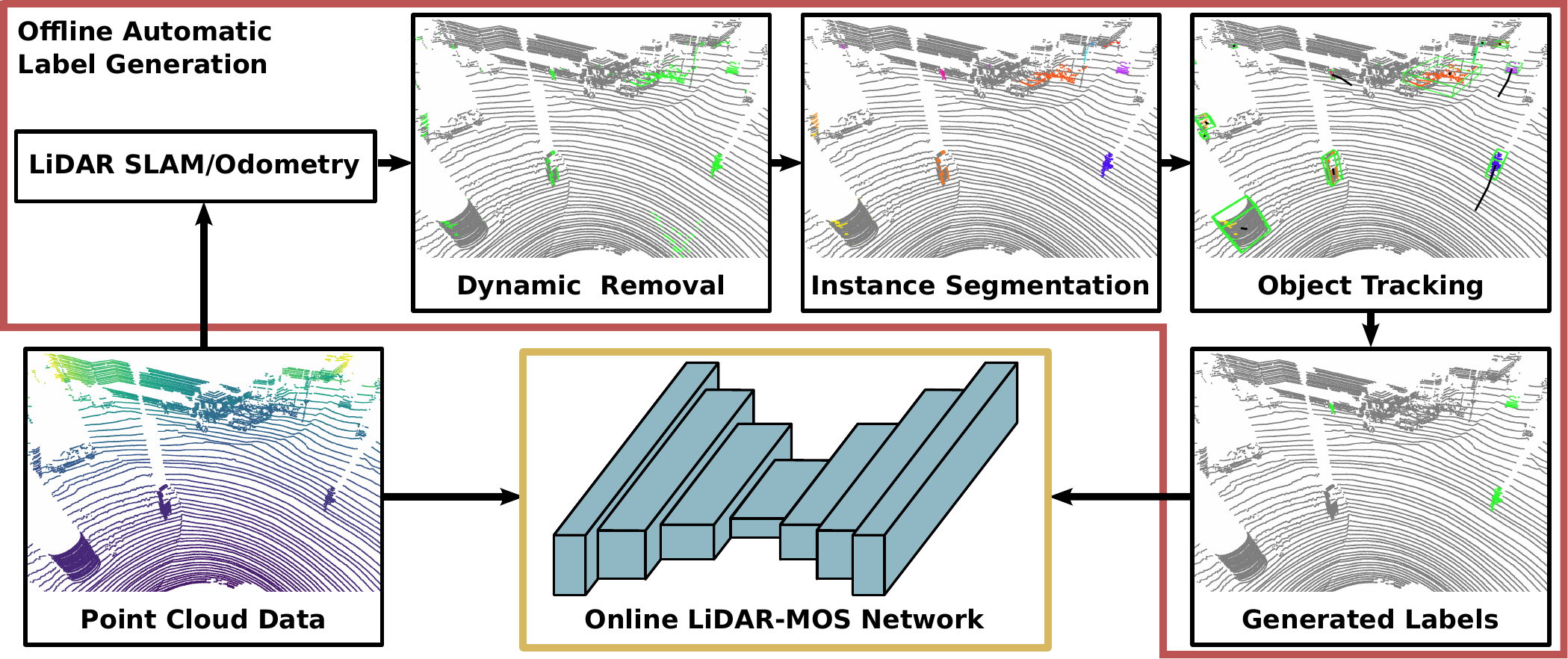}
	\caption{Overview of our method. It uses sequential LiDAR scans as input, and first conducts a LiDAR odometry\,/\,SLAM step to estimate the poses. With the estimated poses, it then applies a map cleaning method to coarsely detect the moving objects (colored in green). A clustering method is then used to extract instances (in different colors) based on the detected moving object proposals. After that, a multi-object tracking method is applied to associate instances (with bounding boxes) and decide the final labels of instances based on the tracked trajectories (colored in black). With the generated labels, we train LMNet that can be later deployed for online LiDAR-MOS.}
	\label{fig:overview}
  \vspace{-0.5cm}
\end{figure*}

	The approach closest to our work is the one by Pfreundschuh~\etalcite{pfreundschuh2021icra}. It generates labels automatically to train a network for moving object-aware SLAM working in small areas,~\eg, a hall or a train station.
	Different from their work, our approach does not require ground truth poses and exploits tracking to distinguish between moving and non-moving objects. Furthermore, our method can be used in large-scale outdoor scenes.

%%%%%%%%%%%%%%%%%%%%%%%%%%%%%%%%%%%%%%%%%%%%%%%%%%%%%%%%%%%%%%%%%%%%%%%%%%%%%%%%
\section{Our Approach}
\label{sec:main}
	We are interested in automatically labeling LiDAR points offline with two classes, \emph{static} or \emph{dynamic}, that can be later used for training a network for online MOS.
	Our approach consists of five serialized modules as shown in~\figref{fig:overview}.
	Our proposed method only uses sequential LiDAR scans as input, and first uses a LiDAR odometry\,/\,SLAM approach to estimate the poses for each scan (see~\secref{sec:slam}). With the estimated poses, we then apply a map cleaning method to coarsely detect moving objects (see~\secref{sec:map-clean}). We then use a clustering method to extract instances based on the detected moving object proposals (see~\secref{sec:cluster}). After that, we apply a multi-object tracking method to associate instances and determine the final labels of instances based on the tracked trajectories (see~\secref{sec:tracking}). Once we have generated the labels offline, we train LMNet that can be later deployed on an autonomous vehicle to perform LiDAR-MOS online in an unseen environment (see~\secref{sec:cnn}).
%  The individual approaches for each step can be replaced easily due to the highly modularized nature of our proposed framework. 
  % Note that our approach is not bound to specific methods which makes it possible to integrate better performing alternatives in the future.
	
%%%%%%%%%%%%%%%%%%%%%%%%
\subsection{LiDAR Odometry}
\label{sec:slam}	
	Instead of exploiting ground-truth poses or information from other sensors, such as RTK-GPS, our method uses only sequential LiDAR scans as input.
	We use an off-the-shelf SLAM approach, SuMa~\cite{behley2018rss}, to estimate the pose of each LiDAR scan, but other systems might be used instead. SuMa exploits a spherical projection of the point cloud and estimates the relative pose between the current LiDAR scan and the maintained world model via projective ICP.
	% Following SuMa~\cite{behley2018rss}, we denote the transformation of a point $\q{p}_A$ in coordinate frame $A$ to a point $\q{p}_B$ in coordinate frame $B$ by $\mq{T}_{BA} \in \RR^{4\times4}$, such that $\q{p}_B = \mq{T}_{BA} \q{p}_A$, where $\q{p}_B, \mq{T}_{BA}, \q{p}_A$ are all in homogeneous coordinates.
	% Let $\m{R}_{BA} \in \mathrm{SO}(3)$ and $\v{t}_{BA} \in \RR^{3}$ denote the corresponding rotational and translational part of the transformation $\mq{T}_{BA}$.
	% We use the notion $C_t$ for the coordinate frame at timestep $t$. Each variable in coordinate frame $C_t$ is associated to the world frame $W$ by a pose $\mq{T}_{WC_t} \in \RR^{4 \times 4}$, transforming the observed point cloud into the world coordinate frame.
	In the following, we denote the estimated absolute pose of a scan at time $t$ by $\mq{T}_t \in \RR^{4 \times 4}$.

	Note that there might be noise in the estimated poses, which may influence the performance of visibility or ray tracing-based methods~\cite{kim2020iros, pfreundschuh2021icra}. With increasing noise, more objects will be detected as moving due to the misalignment caused by inaccurate poses. However, this does not affect our method, since we use the estimated poses for map cleaning and generate only coarse dynamic object proposals, which are later verified and typical false positives are filtered out via tracking.

%%%%%%%%%%%%%%%%%%%%%%%%
\subsection{Coarse Dynamic Object Removal}
\label{sec:map-clean}

	Instead of using semantic information to pre-define movable objects~\cite{chen2019iros}, \eg, cars, pedestrians, we only exploit the sequential temporal-spatial information to coarsely detect the moving objects in a class-agnostic manner. 
	Since the sequential LiDAR data is available for offline label generation, we can detect dynamic objects with an aggregated  map, which is also referred as map cleaning.
	In this work, we use ERASOR~\cite{lim2021ral}, a state-of-the-art map cleaning approach. It first aggregates all local LiDAR points to obtain a single point cloud map $\mathcal{M}$:
	\begin{align}
\mathcal{M}&=\bigcup_{t\in \mathcal{T}}\left\{{\mq{T}_{t} \q{p} | \q{p} \in \mathcal{P}_{t}}\right\} ,
		  %  &= \{\q{p}_i\}^{N_\mathcal{M}}_{i=1},
\end{align}
where $\mathcal{T} = \{1, 2, \dots, N\}$ is the set of timestamps and $N$ is the number of scans. $\mathcal{P}_{t} = \{\q{p}_j\}^{N_{t}}_{j=1} $ is the scan at time $t$ with $N_t$ points. %$N_{\mathcal{M}}$ is the number of points of the aggregated map $\mathcal{M}$. 
The transformations $\mq{T}_t$ are the aforementioned estimated poses from the SLAM method.

Let $\hat{\mathcal{M}}$ be the estimated static map, where the problem of map cleaning is defined as follows:
\begin{align}
\hat{\mathcal{M}}&=\mathcal{M}-\bigcup_{t\in \mathcal{T} }{\hat{\mathcal{M}}_{\text{dyn},t}} ,
\end{align}  
where $\mathcal{\hat{\mathcal{M}}}_{\text{dyn}, t}$ refers to the estimated dynamic points at timestamp $t$. In this work, we are interested in the set $\mathcal{\hat{\mathcal{M}}}_{\text{dyn}, t}$ of moving objects instead of the static map~$\hat{\mathcal{M}}$.

ERASOR determines dynamic points by checking the discrepancy between transformed points $\mathcal{P}'_t = \left\{{\mq{T}_{t} \q{p} | \q{p} \in \mathcal{P}_{t}}\right\}$ and $\mathcal{M}$, \ie, if an object in the map $\mathcal{M}$ can be found in the same position of $\mathcal{P}'_t$. Instead of checking the whole query scan $\mathcal{P}'_t$ at the same time $t$, it divides the volume of interest into small cells over azimuthal and radial directions. For each cell, it calculates the pseudo occupancy by subtracting the maximum and minimum height of the points inside the cell, and then labels the cell as dynamic if the ratio of pseudo occupancy between the corresponding pair of cells in the query $\mathcal{P}'_t$ and map $\mathcal{M}$ is larger than a threshold. In the end, it reverts the ground points as static in the labeled dynamic cells. For more details, we refer to the original paper~\cite{lim2021ral}.
	
Even though such map cleaning methods can distinguish moving objects from the static map, we observe a substantial number of false-positive detections, possibly caused by noisy points or inaccuracies in the estimated poses. Since the goal of ERASOR is to obtain a static map, it is reasonable for map cleaning methods to be more aggressive and remove some points to guarantee a clean map. Moreover, exploiting both, past and future data sequentially, there are also redundant observations. Thus, it may not influence the mapping results when cleaning methods remove more points that belong to static objects.

However, the objective of MOS training data generation is different from that of map cleaning. Thus, we aim to accurately separate actual moving objects, \eg, driving cars,   from static or non-moving objects, \eg, buildings, parked cars.

%%%%%%%%%%%%%%%%%%%%%%%%
\subsection{Class-agnostic Instance Segmentation}
\label{sec:cluster}	

To compute accurate segments of moving objects, we apply instance segmentation on the dynamic proposals $\mathcal{\hat{\mathcal{M}}}_{\text{dyn}, t}$ provided by the map cleaning method. The goal of a segmentation $\mathcal{S}$ is to partition the point cloud into disjoint subsets:
\begin{align}
\mathcal{S}&=\bigcup_{k \in \{1, \dots, N_S\}} \mathcal{S}_k ,
\end{align} 
where $\mathcal{S}_k \subset \mathcal{\hat{\mathcal{M}}}_{\text{dyn}, t}$ is a segment, and $N_S$ denotes the number of segments.
Every point $\q{p} \in \hat{\mathcal{M}}_{\text{dyn}, t}$ exists in one and only one segment $\mathcal{S}_k$, \ie, 
$\mathcal{S}_i \cap \mathcal{S}_j = \emptyset, \forall\, i \neq j$.

There are many methods for class-agnostic segmentation~\cite{behley2013iros,bogoslavskyi2016iros,campello2015tkdd}. 
In this paper, we choose HDBSCAN~\cite{campello2015tkdd}, which is an extension of DBSCAN~\cite{campello2013dbscan}. DBSCAN is density-based and provides a clustering hierarchy from which a simplified tree of significant clusters can be constructed. HDBSCAN performs DBSCAN over varying density thresholds $\epsilon$ and integrates the result yielding a clustering that gives the best score over $\epsilon$. This allows HDBSCAN to find clusters of varying densities (unlike DBSCAN), and to be more robust to parameter selection. % We refer to the original papers~\cite{campello2015tkdd, campello2013dbscan} for further details.

Once we obtain the final clustering results using HDBSCAN, we generate a bounding box $\v{b}_k = (\v{c}_k, \theta, l, w, h, s)$ for each segment $\mathcal{S}_k$, including the center coordinates~$\v{c}_k \in \RR^3$, the length $l$, width $w$ and height $h$, heading angle $\theta$, and confidence score $s$. We filter out segments that are too small or too large, \ie, segments with less than $N_\text{min}$ points or with a maximum side length of the bounding box larger than a threshold $T_\text{size}$ and get the final set of instances $\mathcal{B}~=~\{\v{b}_m\}_{m=1}^{N_B}$ with $N_B \le N_S$.

%%%%%%%%%%%%%%%%%%%%%%%%
\subsection{Multiple Dynamic Object Tracking}
\label{sec:tracking}	

After clustering, there can still be static objects inside the set of instances $\mathcal{B}$. To verify the real dynamic objects, we use a multi-object tracking method proposed by Weng \etal~\cite{weng2020iros} to obtain trajectories of object instances and determine the label for each tracked instance based on its movement.

To this end, we use multiple extended Kalman filters to track instance bounding boxes. For the motion model, we first compensate the ego-motion using the poses estimated by the SLAM system. Then, for each instance, we apply a constant velocity model as the state prediction. For the observation model, we find associations between instances in consecutive scans. We compute a cost matrix $\m{C}\in \RR^{N_B^t\times N_B^{t-1}}$ between all $N_B^t$ instances in the LiDAR scan at timestamp $t$ and all $N_B^{t-1}$ instances still tracked in the previous scan at timestamp $t-1$, by means of the similarity. The association problem can be formulated as a bipartite graph matching problem that can be solved using the Hungarian method~\cite{kuhn1955nrlq}, an optimal assignment method, to determine the pairs of associations between currently detected and previously tracked instances.

To compute the instance similarity, we take three different geometric features into account, the center distance, the overlapping bounding box volumes, and the change of the volume between each pair of instances based on their bounding boxes. Each entry of the cost matrix $\m{C}$ is calculated as the association cost by a linear combination of these three features between a new instance $i$ and the prediction of a previous tracked instance $j$:
\begin{align}
  \m{C}_{i,j} &= \alpha_d \, c_d + \alpha_o \, c_o  + \alpha_v \, c_v, \\
  c_d &= \lVert\v{c}_i-{\v{c}}_j\lVert_2,  \\
  c_o &= 1-\text{IoU}(\v{b}_i,{\v{b}}_j) , \\
  c_v &= 1- \frac{\text{min}(\v{v}_i,{\v{v}}_j)}{\text{max}(\v{v}_i,{\v{v}}_j)} ,
  \label{eq:cost}
\end{align}
where $c_d, c_o, c_v~\in~\RR$ are the cost for the center distance, overlapping volume, and the change of the volume, and $\alpha_d, \alpha_o, \alpha_v~\in~\RR$ are corresponding importance weights. The center of the bounding box  $\v{b}_k$ is denoted as $\v{c}_k$ and $\lVert \cdot \lVert_2$ is the Euclidean distance. The intersection over union $\text{IoU}\left(\cdot\right)$ between two instance bounding boxes is used to account for the volume overlap with $\v{v}_k = l\times w \times h$ representing the volume of the bounding box $\v{b}_k$.
	
There are multiple challenging cases during tracking. For example, newly detected objects might not be tracked in the previous scans, a tracked instance might leave the scene, or a tracked instance might be occluded or missed. 
To avoid wrong associations, we add a new EKF in case of newly detected objects by determining $\text{Flag}_{\text{add}}$:
\begin{align}
  \text{Flag}_{\text{add}} = (c_d > T_d) \lor (c_o > T_o) \lor (c_v > T_v), 
\end{align}
where $T_d, T_o, T_v$ are the thresholds for the three instance features respectively determined based on the validation data. 
We store the deactivated trajectories for a fixed number of timestamps~$n_\text{old}$. 
We associate or re-identify instances if the similarity between a newly detected object and the still tracked or deactivated one is high, resulting in \mbox{$\text{Flag}_{\text{add}} = \mathit{false}$}.
For re-identification, the motion model is applied continuously also to the deactivated targets (not matched with any newly detected instance) to estimate their position in case of occlusions or missed detections. 

After tracking, we label an instance as moving if its accumulated trajectory is larger than its maximum side length of the associated bounding box. We label the instance point-wise, which means all points inside the bounding box will be labeled as moving once we determine the instance as moving. In case of misdetection or occlusion, we can still generate temporally consistent labels by re-identifying instances using the EKFs.

%%%%%%%%%%%%%%%%%%%%%%%%
\subsection{Training a Neural Network for Online LiDAR-MOS}
\label{sec:cnn}
	
After running our pipeline, we obtain binary labels for all points. Based on that, we can train our LiDAR-MOS network LMNet~\cite{chen2021ral}, which can be later deployed for online LiDAR-MOS in unseen environments. LMNet exploits sequential range images from a LiDAR sensor as an intermediate representation combined with a typical encoder-decoder CNN and runs faster than the frame rate of the sensor.
By using residual images, our network obtains temporal information and can differentiate between moving and static objects and achieves state-of-the-art performance on the LiDAR-MOS task.
For more details, we refer to the LMNet paper~\cite{chen2021ral}.

Instead of using the manual labels provided by our LiDAR-MOS benchmark, we train our LMNet with the automatically generated labels by our proposed offline pipeline. Since our method can generate labels automatically for sequential LiDAR data, we can train LMNet with more data than that provided by the LiDAR-MOS benchmark~\cite{chen2021ral}, which further boosts the performance of our network for online LiDAR-MOS.
Furthermore, the proposed auto-labeling method can also generate labels for LiDAR data from other datasets, which were collected from different environments. Using these additional generated labels, our LMNet can better generalize in different environments.
% We will show the improved LiDAR-MOS performance in the following section.

%%%%%%%%%%%%%%%%%%%%%%%%
\subsection{Parameters and Implementation Details}
\label{sec:implementation}
Our method consists of individual modules that are carried out sequentially. We use SuMa~\cite{behley2018rss} with the default parameters as the LiDAR SLAM approach. We then use ERASOR~\cite{lim2021ral} with the default parameters to coarsely detect the dynamic objects. Based on the dynamic object proposals, we apply HDBSCAN~\cite{campello2015tkdd} with multiple density thresholds~$\epsilon=\{2.0, 1.0, 0.5, 0.25\}$ to generate dynamic object instances. After filtering out instances with less than $N_\text{min}\,=\,5$ points or with a maximum side length larger than $T_\text{size}\,=\,20\,\mathrm{m}$, we track the remaining instances using a multi-object tracking method~\cite{weng2020iros}. We use the same importance weights for three different association terms, \ie, $\alpha_d\,=\,\alpha_o\,=\,\alpha_v\,=\,1$, and keep track of deactivated instances for $n_\text{old}\,=\,5$ timestamps. The thresholds are $T_d\,=\,2\,\mathrm{m}$, $T_o\,=\,0.95$, and $T_v\,=\,0.7$.
For training LMNet~\cite{chen2021ral}, we use the default parameters with $8$ residual images without semantic information and train for $150$ epochs.

%%%%%%%%%%%%%%%%%%%%%%%%%%%%%%%%%%%%%%%%%%%%%%%%%%%%%%%%%%%%%%%%%%%%%%%%%%%%%%%%
\section{Experimental Evaluation}
\label{sec:exp}

\begin{figure*}[t]
%    \vspace{0.2cm}
	\centering
	\subfigure[Octomap-based~\cite{arora2021ecmr}]{\includegraphics[width=0.324\linewidth]{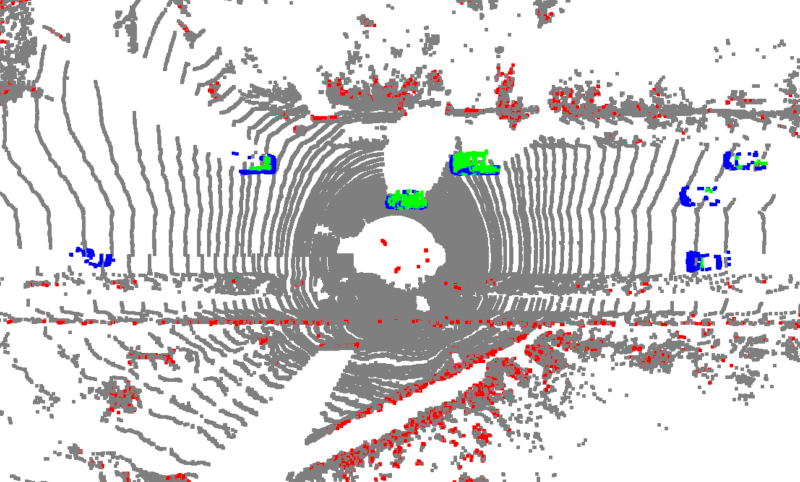}}
	\subfigure[Removert~\cite{kim2020iros}]{\includegraphics[width=0.324\linewidth]{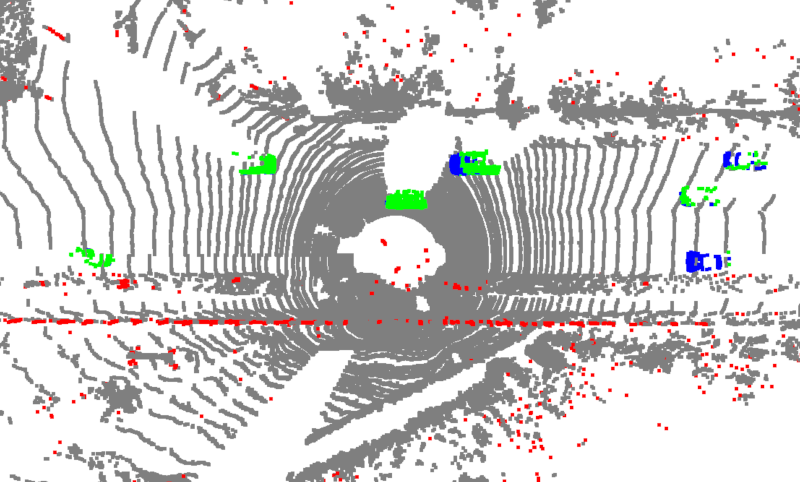}}
	\subfigure[Erasor~\cite{lim2021ral}]{\includegraphics[width=0.324\linewidth]{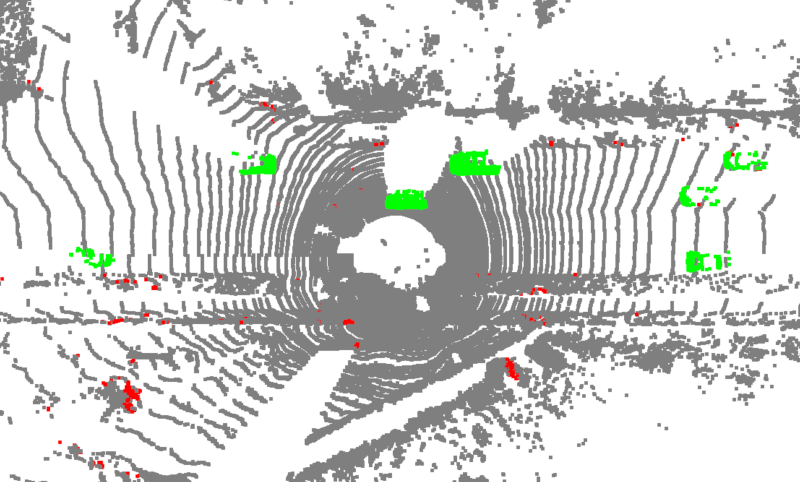}}
	\subfigure[Yoon's~\cite{yoon2019cvr}]{\includegraphics[width=0.324\linewidth]{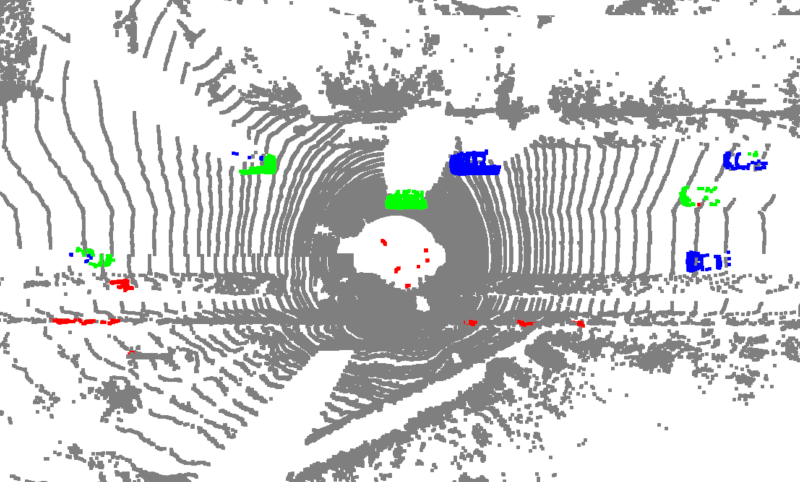}}
	\subfigure[Pfreundschuh's~\cite{pfreundschuh2021icra}]{\includegraphics[width=0.324\linewidth]{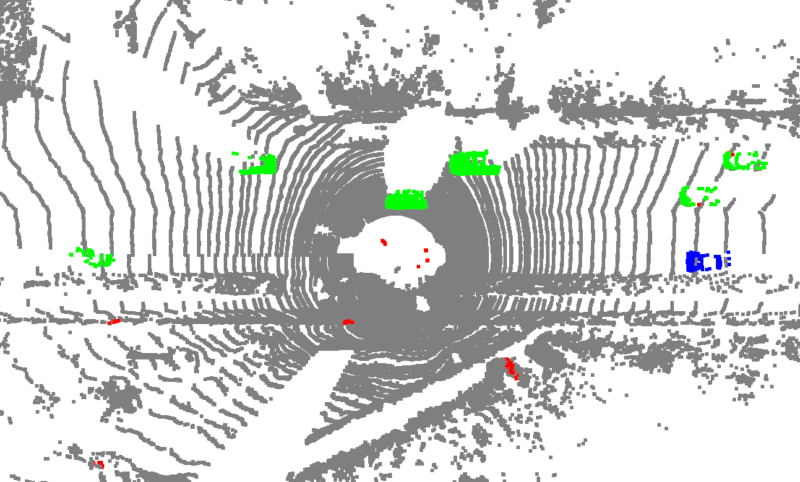}}
	\subfigure[Ours]{\includegraphics[width=0.324\linewidth]{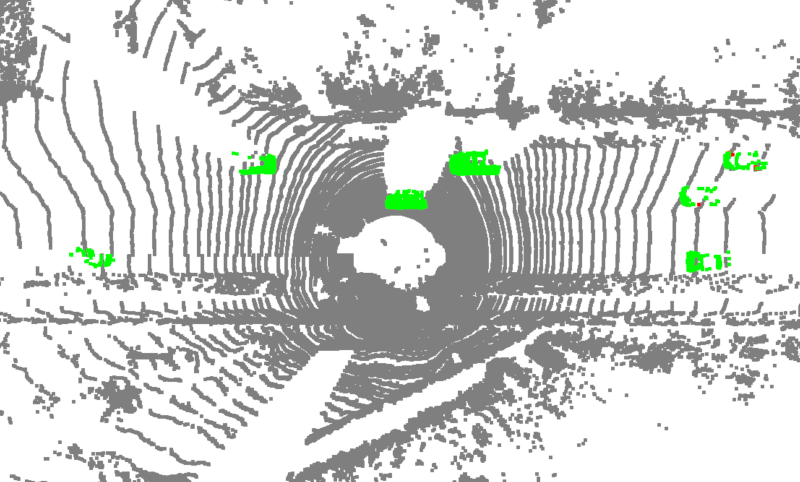}}
	\caption{Qualitative results of (a) Octomap-based~\cite{arora2021ecmr}, (b) Removert~\cite{kim2020iros}, (c) Erasor~\cite{lim2021ral}, (d) Yoon's~\cite{yoon2019cvr}, (e) Pfreundschuh's~\cite{pfreundschuh2021icra}, and (f) ours. The green points represent the true dynamics (TD), red points are false dynamics (FD), blue points are false statics (FS), and the gray background are true statics (TS).}
	\vspace{-0.5cm}
	\label{fig:auto-label-qualitative}
\end{figure*}
	
%The main focus of this work is to train a deep learning-based approach to segment moving objects in sequential LiDAR data exploiting automatically generated labels.
	
We present our experiments to illustrate the capabilities of our method. 
They furthermore support our key claims, that:
(i) Our approach generates better labels for moving object segmentation using only 3D LiDAR scans;
(ii) Based on our automatically generated labels, the network achieves a similar performance in LiDAR-MOS compared to the one trained with manual labels, and a better performance with additional automatically generated training data; 
(iii) Our method generates LiDAR-MOS labels for different LiDAR data collected from different environments.
	
%%%%%%%%%%%%%%%%%%%%%%%%
\subsection{Experimental Setup}
\label{subsec:setup}
For quantifying the LiDAR-MOS performance, we use the intersection-over-union (IoU) metric over moving objects, which is given by
\begin{align}
	\text{IoU}_{\text{MOS}} & = \frac{\text{TD}}{\text{TD} + \text{FD} + \text{FS}}, \label{eq:miou}
\end{align}
where $\text{TD}$, $\text{FD}$, and $\text{FS}$ correspond to the number of true dynamics, false dynamics, and false statics points.
	
We evaluate our method on four different datasets. The first one is the LiDAR-MOS benchmark~\cite{chen2021ral} of SemanticKITTI~\cite{behley2021ijrr}, which separates all classes into moving and static. It contains 22 sequences, from 00 to 07 and 09 to 10 for training with ground truth labels, 08 for validation, and 11 to 21 for testing. In this work, we first evaluate the automatic label generation methods on the training data in~\secref{subsec:auto-label}. After generating the labels automatically, we re-train the LMNet with the generated labels and test the re-trained model on the hidden test set. Since there is more unlabeled LiDAR data available in the original KITTI dataset~\cite{geiger2013ijrr}, we generate more labels automatically and train the network with extra data. We also test the further trained model on the hidden test set, see~\secref{subsec:lidar-mos}. 
To evaluate the generalization ability of the proposed method, we also test our approach on three other datasets, Apollo~\cite{lu2019cvpr}, MulRan~\cite{kim2020icra}, and IPB-Car~\cite{chen2021auro, chen2020iros}, shown in~\secref{subsec:generalization}. 
%The KITTI dataset was collected in Germany in 2011, while the Apollo dataset was collected in the U.S. in 2018, MulRan in Korea in 2019, and IPB-Car in Germany in 2020 (shown in~\secref{subsec:generalization}). 
KITTI and Apollo have been recorded with Velodyne-64 LiDAR scanners, while IPB-Car and MulRan offer LiDAR data from Ouster sensors.

%%%%%%%%%%%%%%%%%%%%%%%%
\subsection{Automatic Data Labelling Results}
\label{subsec:auto-label}
The experiment in this section supports our claim that our approach generates better labels for moving object segmentation using only 3D LiDAR scans than baseline methods. We test multiple methods that can distinguish moving and non-moving objects on 3D LiDAR data, as shown in~\tabref{tab:auto-label}.	We compare our method to map cleaning-based methods like Removert~\cite{kim2020iros}, ERASOR~\cite{lim2021ral}, and an Octomap-based~\cite{arora2021ecmr} method. Those methods are open-source and have been evaluated on the KITTI dataset. Therefore, we directly use the provided implementation with the default parameters.	We also re-implement two other methods whose source code is not publicly available. The first one is proposed by Yoon~\etalcite{yoon2019cvr} and detects dynamic objects in LiDAR scans exploiting geometric heuristics, including residual and region growth, named ``Yoon's''. The other one is the label generation method proposed by Pfreundschuh~\etalcite{pfreundschuh2021icra}, which uses range images for visibility checking and later uses clustering to verify candidates by voting, named ``Pfreundschuh's''. We implement that method using the range image-based visibility checking part from Removert, the HDBSCAN for clustering, and the same voting as used in the original paper.	
Our method is denoted as ``Ours''. 
%We also provide the voting-based version which uses the same voting strategy as introduced by~Pfreundschuh~\etalcite{pfreundschuh2021icra} called \emph{Ours-voting}.	
All methods only use geometric information without any learning or semantic information, which makes it possible to generate labels fully unsupervised. 

As shown in~\tabref{tab:auto-label}, our method outperforms other baseline methods on generating LiDAR-MOS labels in terms of IoU over moving objects, which is in line with the qualitative results shown in~\figref{fig:auto-label-qualitative}.
As can be seen, the map-cleaning-based methods, Octomap-based, Removert, and ERASOR, show many false dynamics (colored in red), while other label generation methods, ``Yoon's'' and ``Pfreundschuh's'', often miss moving objects leading to many false static points (colored in blue). In comparison, our method detects moving objects more accurately.
	
\begin{table}[t]
%    \vspace{0.2cm}
	\caption{Evaluation on automatic label generation.}
	\centering

	\begin{tabular}{L{6cm}r}
		\toprule
		                                   & $\text{IoU}_{\text{MOS}}$  \\
		\midrule
		Octomap-based \cite{arora2021ecmr}              	   & 13.6 \\
		Removert \cite{kim2020iros}             		       & 15.7 \\
		ERASOR \cite{lim2021ral}              		       & 19.1 \\
		\midrule
		Pfreundschuh's \cite{pfreundschuh2021icra}                    & 34.5  \\
		Yoon's \cite{yoon2019cvr}                            & 15.1 \\
		\midrule
%		Ours-voting            			   & 46.5 \\
		Ours                               & $\mathbf{74.2}$ \\
		\bottomrule
	\end{tabular}
	\vspace{-0.5cm}
	\label{tab:auto-label}
\end{table}

\begin{table}[t]
	%    \vspace{0.2cm}
		\caption{Online LiDAR-MOS Performance on Benchmark.}
		\centering
	
		\begin{tabular}{L{6cm}r}
			\toprule
																				 & $\text{IoU}_{\text{MOS}}$  \\
			\midrule
			SalsaNext \cite{cortinhal2020iv} (movable classes)		   & 4.4  \\
			SceneFlow \cite{liu2019cvpr}     				   & 4.8  \\
			SqSequence \cite{shi2020cvpr}             		   & 43.2 \\
			KPConv \cite{thomas2019iccv}               		       & 60.9 \\
			\midrule
			LMNet+Manual					   & 58.3 \\
			LMNet+Ours            			   & 54.3 \\
			LMNet+Ours+Extra                   & $\mathbf{62.3}$ \\
			\bottomrule
		\end{tabular}
		\vspace{-0.5cm}
		\label{tab:benchmark}
	\end{table}

%%%%%%%%%%%%%%%%%%%%%%%%
\subsection{MOS Performance using Auto-generated Labels}
\label{subsec:lidar-mos}
In the second experiment, we re-train LMNet with the automatically generated labels and evaluate the model on the hidden test set of the LiDAR-MOS benchmark, as shown in~\tabref{tab:benchmark}. We compare the network trained on manual ground truth labels (LMNet+Manual), with the network re-trained on the labels generated by the proposed method (LMNet+Ours). As can be seen, LMNet+Ours achieves a similar IoU on the benchmark as LMNet+Manual. We also show the results of LMNet re-trained with extra automatically generated labels (LMNet+Ours+Extra). In this experiment, we use the road sequences of KITTI raw dataset~\cite{geiger2013ijrr}, generate additional labels using the proposed method, and re-train LMNet with labels automatically generated on both KITTI odometry and road data. Boosted with extra data, we see that LMNet+Ours+Extra outperforms LMNet+Manual trained with manual labels, and is the best performing strategy.
 
We additionally compare our method to other online LiDAR-MOS methods that only use the current and past observations as required for real-world self-driving applications. For example, one uses all movable objects as dynamic depending on the semantics from a semantic segmentation network, such as SalsaNext~\cite{cortinhal2020iv}, or multi-scan networks to learn to distinguish moving and non-moving object classes, such as SqSequence~\cite{shi2020cvpr} or KPConv~\cite{thomas2019iccv}, or using the flow vectors with a threshold to determine the moving objects, such as SceneFlow~\cite{liu2019cvpr}. Our method also outperforms all these baselines by a large margin. %trained with automatically generated labels.
	
\begin{figure}[t]
	\centering
	\includegraphics[width=0.8\linewidth]{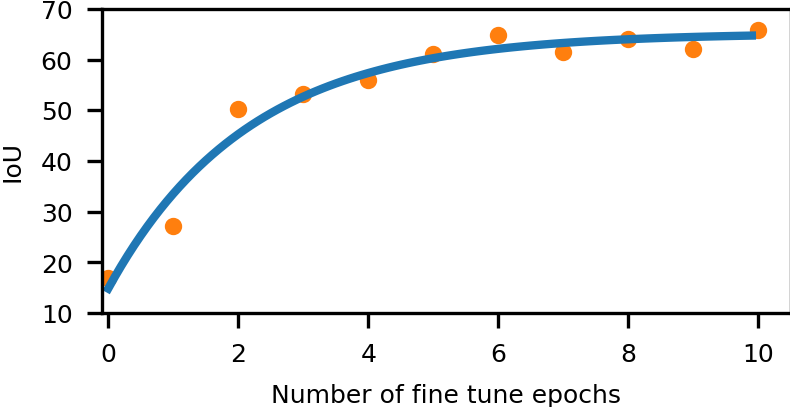}
	\caption{MOS performance \vs the number of epochs for fine-tuning on Apollo data.
}
	\label{fig:epochs}
\end{figure}

\begin{table}[t]
%    \vspace{0.2cm}
	\caption{Online LiDAR-MOS Performance on Apollo.}
	\centering

	\begin{tabular}{L{6cm}r}
		\toprule
		                                   & $\text{IoU}_{\text{MOS}}$  \\
		\midrule
		LMNet+Manual (trained on KITTI)	   & 16.9 \\
		LMNet+Ours            			   & 45.7 \\
		LMNet+Ours+Fine-Tuned                   & $\mathbf{65.9}$ \\
		\bottomrule
	\end{tabular}
	\vspace{-0.3cm}
	\label{tab:apollo}
\end{table}

%%%%%%%%%%%%%%%%%%%%%%%%
\subsection{Generalization Capabilities}
\label{subsec:generalization}
The third experiment investigates the generalization capabilities of our approach, and it supports our last claim that our method can generate LiDAR-MOS labels for different LiDAR data collected from different environments. To this end, we test our method on three more datasets, Apollo~\cite{lu2019cvpr}, MulRan~\cite{kim2020icra}, and IPB-Car~\cite{chen2020iros, chen2021auro}. To quantitatively evaluate the generalization capabilities of the proposed method, we manually label a small test set on the Apollo-ColumbiaPark-MapData~\cite{lu2019cvpr}, sequence 2 frame 22300-24300 and sequence 3 frame 3100-3600, which contain many dynamic objects. We use the same LMNet and compare three different setups: the pre-trained model with SemanticKITTI LiDAR-MOS manual labels  (LMNet+Manual trained on KITTI), the re-trained model with automatically generated labels on Apollo-ColumbiaPark-MapData sequence 1 frame 6500-7500 and sequence 4 frame 1500-3100 (LMNet+Ours), and the pre-trained model with automatically generated labels on KITTI fine-tuned with automatically generated labels on the Apollo dataset (LMNet+Ours+Tuned). 

As shown in~\tabref{tab:apollo}, the model pre-trained with SemanticKITTI LiDAR-MOS manual labels cannot generalize very well to different environments. The re-trained model with our approach achieves a better performance, even when using only a small set of training data, in this case only 2600 LiDAR scans. If we fine-tune the pre-trained model with our automatically generated labels, \ie, using automatic generated labels from both KITTI and Apollo, it performs best with only a few fine-tuning epochs (see~\tabref{tab:apollo} and~\figref{fig:epochs}). 

\begin{figure*}[t]
%    \vspace{0.2cm}
	\centering
	\subfigure[Apollo raw]{\includegraphics[width=0.325\linewidth]{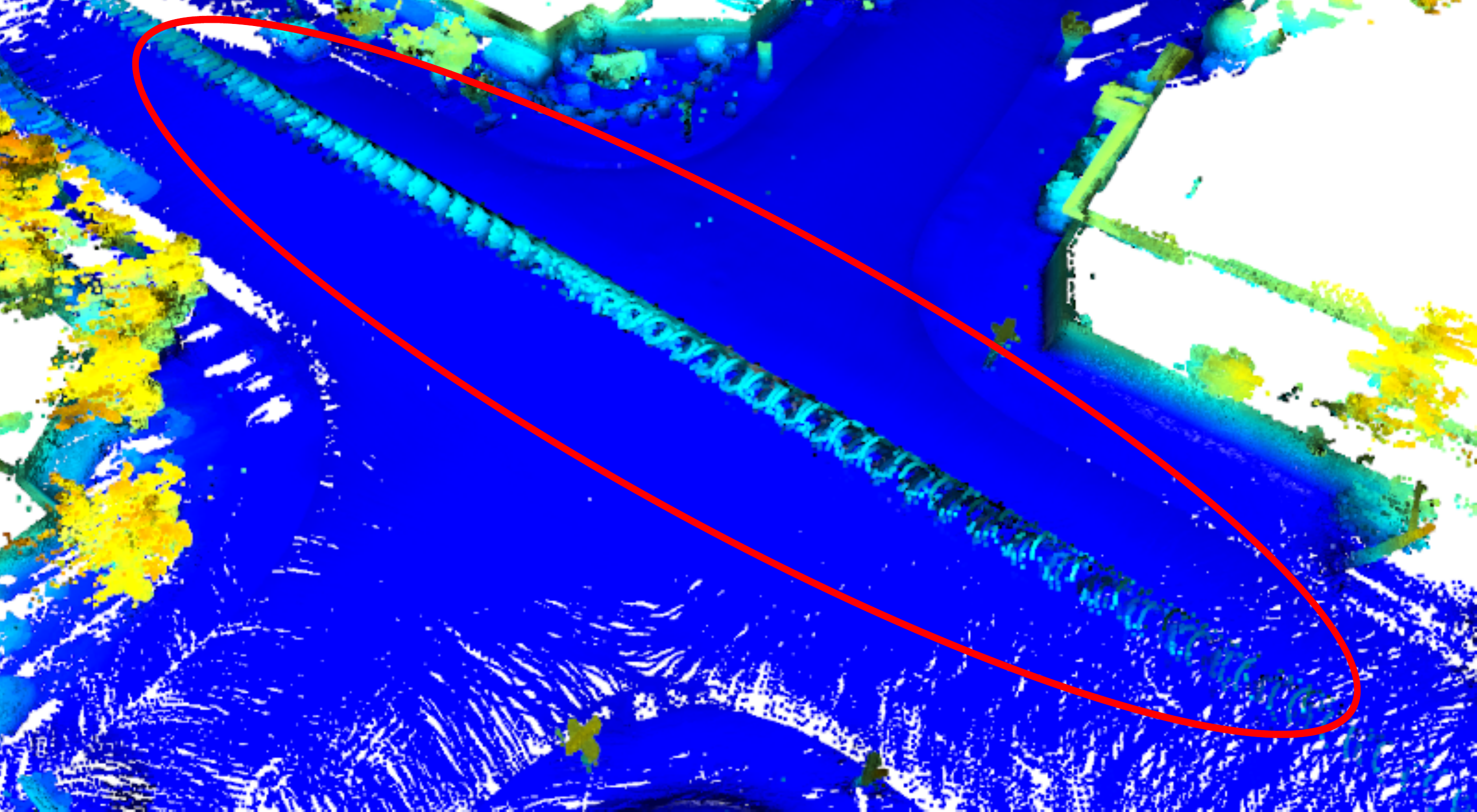}}
	\subfigure[IPB-Car raw]{\includegraphics[width=0.325\linewidth]{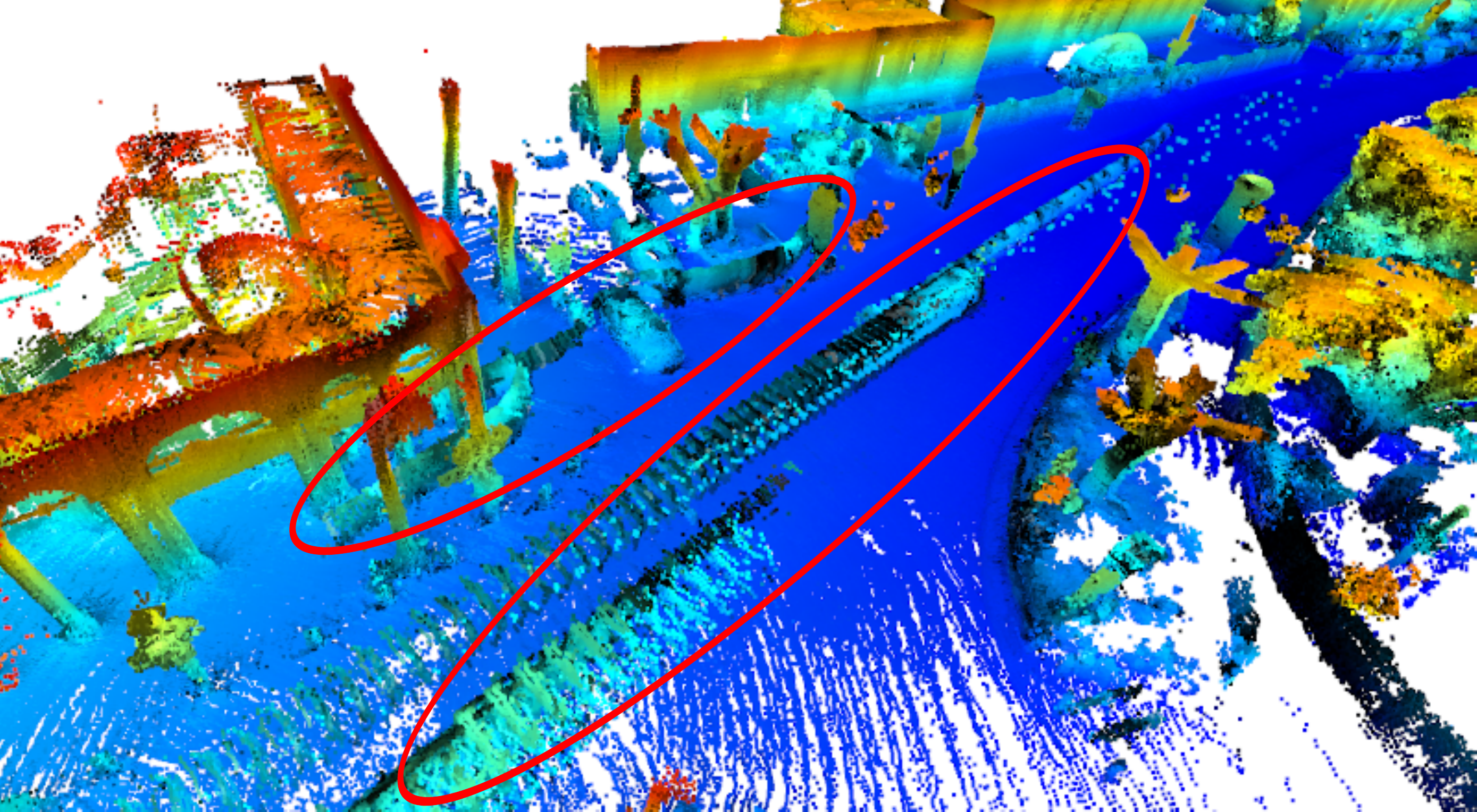}}
	\subfigure[MulRan raw]{\includegraphics[width=0.325\linewidth]{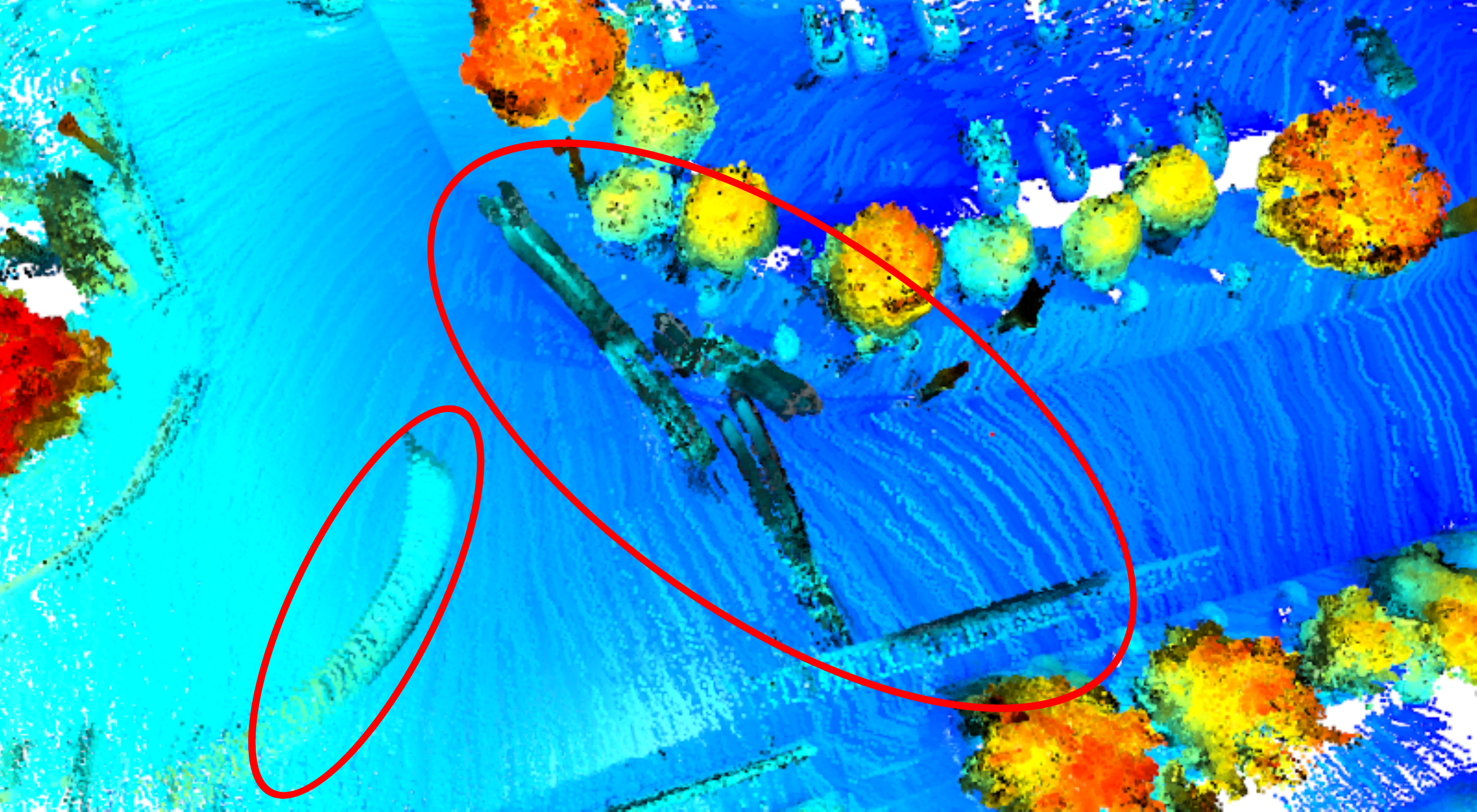}}
	\subfigure[Apollo clean]{\includegraphics[width=0.325\linewidth]{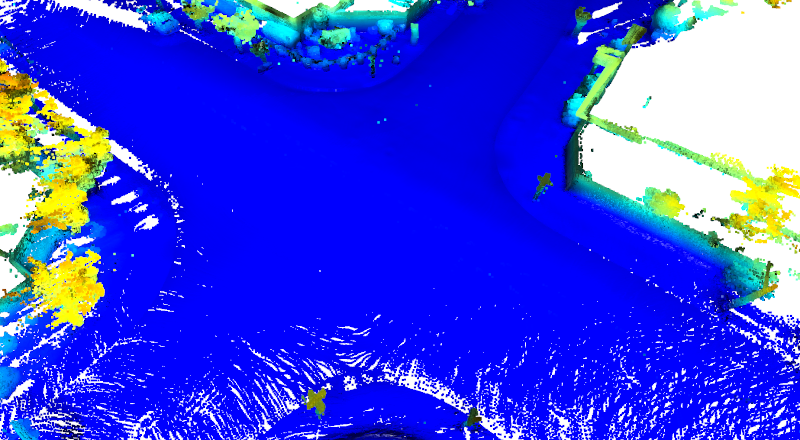}}
	\subfigure[IPB-Car clean]{\includegraphics[width=0.325\linewidth]{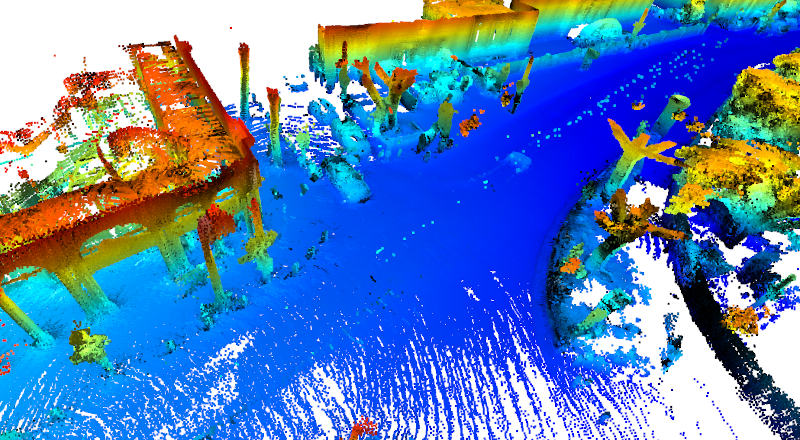}}
	\subfigure[MulRan clean]{\includegraphics[width=0.325\linewidth]{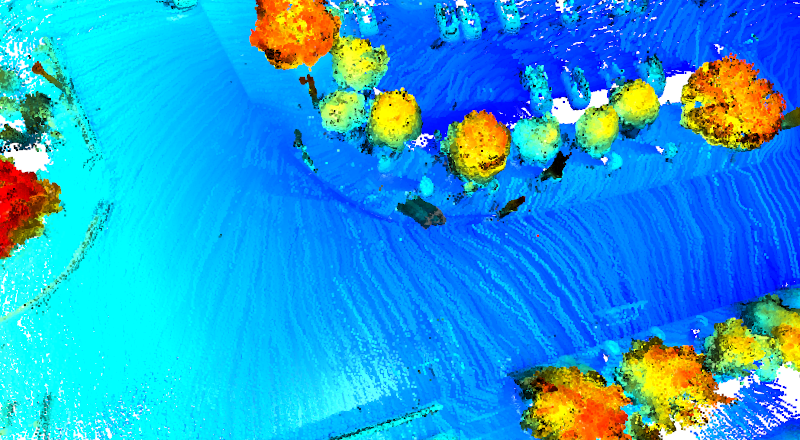}}
	\caption{Map cleaning results on three different datasets. The color from blue to red represents the height from low to high. Artifacts by moving objects are effectively removed by our method.}
	\label{fig:clean-map}
    \vspace{-0.5cm}
\end{figure*}

We also test our method on the MulRan and IPB-Car datasets. Since there are no manual labels available, we only show qualitative mapping results of Apollo, IPB-Car, and MulRan datasets in~\figref{fig:clean-map}.
	As can be seen, using the automatically generated labels, we can effectively remove dynamics during mapping (shown in the upper row) and obtain comparably clean maps (shown in the bottom row).

%%%%%%%%%%%%%%%%%%%%%%%%%%%%%%%%%%%%%%%%%%%%%%%%%%%%%%%%%%%%%%%%%%%%%%%%%%%%%%%%
\section{Conclusion}
\label{sec:conclusion}

	In this paper, we presented a novel approach to automatically generate labels on point clouds for moving object segmentation in 3D LiDAR data.	
Our method combines LiDAR odometry, map cleaning, instance segmentation, and multi-object tracking to determine labels for moving and non-moving objects in LiDAR data offline. Based on the labels generated automatically by our approach, we re-train a network to achieve an efficient online LiDAR-MOS system.
	We provided comparisons to other existing techniques and supported all our claims made in this paper. The experiments suggest that our method achieves a solid performance on LiDAR-MOS label generation and substantially boosts the online performance by additional automatically generating labels.
	We evaluated our method on four different datasets, showing strong generalization capabilities for our approach.

%%%%%%%%%%%%%%%%%%%%%%%%%%%%%%%%%%%%%%%%%%%%%%%%%%%%%%%%%%%%%%%%%%%%%%%%%%%%%%%%
% Only if applicable
%\section*{Acknowledgments}
%We thank XXX for fruitful discussions and for \dots
	
\bibliographystyle{plain_abbrv}

% All new citations should go to new.bib. The file glorified.bib should go
% be the one from the ipb server. After paper or related work has been
% written merge the entries from new.bib to glorified.bib ON THE SERVER,
% replace the glorified.bib in this repository and empty the new.bib
\bibliography{glorified,new}
	
\end{document}

%% file: stachnisslab-latex.tex
\usepackage{graphics}           
\usepackage{times}              
\usepackage{amsmath}            
\usepackage{amssymb}            
\usepackage{graphicx}
\usepackage{algorithm}
\usepackage[noend]{algpseudocode}
\usepackage{booktabs}
\usepackage{color}
\definecolor{instructioncolor}{rgb}{.5,.5,.5}

\usepackage[font=small]{caption}

\def\secref#1{Sec.~\ref{#1}}
\def\figref#1{Fig.~\ref{#1}}
\def\tabref#1{Tab.~\ref{#1}}
\def\eqref#1{Eq.~(\ref{#1})}

\makeatletter
\usepackage{xspace}
\DeclareRobustCommand\onedot{\futurelet\@let@token\@onedot}
\def\@onedot{\ifx\@let@token.\else.\null\fi\xspace}
\def\eg{e.g\onedot} 
\def\ie{i.e\onedot} 
 
 \def\vs{vs\onedot}
 
\def\etal{{et al}\onedot}
\makeatother

\def\etalcite#1{\etal~\cite{#1}}

\usepackage{array}
\newcolumntype{L}[1]{>{\raggedright\let\newline\\\arraybackslash\hspace{0pt}}m{#1}}
\newcolumntype{C}[1]{>{\centering\let\newline\\\arraybackslash\hspace{0pt}}m{#1}}
\newcolumntype{R}[1]{>{\raggedleft\let\newline\\\arraybackslash\hspace{0pt}}m{#1}}

%% file: stachnisslab-math.tex
\newcommand{\RR}{\mathbb{R}}

\renewcommand{\b}[1]{\mbox{\boldmath$#1$}}

\renewcommand{\v}[1]{{\b #1}} 

\newcommand{\m}[1]{{\mbox{{\sffamily\slshape{#1\/}}}}}

\newcommand{\q}[1]{{{\bf #1}}}

\newcommand{\mq}[1]{{\mbox{{\sffamily{#1}}}}}